\documentclass[11pt,a4paper]{article}
\usepackage{authblk}
\usepackage[hyperref]{acl2021}
\usepackage{times}
\usepackage{latexsym}

\usepackage{url}

\usepackage{microtype}

\usepackage{graphicx}
\usepackage{amsmath}
\usepackage{amssymb} 
\usepackage{booktabs}
\usepackage{xcolor}
\usepackage{enumitem}
\usepackage{float}
\usepackage{xspace}
\usepackage{float}

\aclfinalcopy 

\definecolor{ao}{rgb}{0.0, 0.5, 0.0}
\definecolor{tb}{cmyk}{0, 0.7808, 0.4429, 0.1412}

\providecommand{\GBERT}{Ger-BERT\xspace}

\title{Investigating label suggestions for opinion mining in German Covid-19 social media} 

\author[1]{\textbf{Tilman Beck}}
\author[1]{\textbf{Ji-Ung Lee}}
\author[2]{\textbf{Christina Viehmann}}
\author[2]{\textbf{Marcus Maurer}}
\author[2]{\textbf{Oliver Quiring}}
\author[1]{\textbf{Iryna Gurevych}}
\affil[1]{Ubiquitous Knowledge Processing Lab, Technical University of Darmstadt}
\affil[2]{Institut f{\"u}r Publizistik, Johannes Gutenberg-University Mainz}
\affil[ ]{\url{www.ukp.tu-darmstadt.de} \protect\\
\url{www.ifp.uni-mainz.de}}

\date{}

\begin{document}
\maketitle
\begin{abstract}

This work investigates the use of interactively updated label suggestions to improve upon the efficiency of gathering annotations on the task of opinion mining in German Covid-19 social media data.
We develop guidelines to conduct a controlled annotation study with social science students and find that suggestions from a model trained on a small, expert-annotated dataset already lead to a substantial improvement -- in terms of inter-annotator agreement (+.14 Fleiss' $\kappa$) and annotation quality -- compared to students that do not receive any label suggestions.
We further find that label suggestions from interactively trained models do not lead to an improvement over suggestions from a static model. Nonetheless, our analysis of suggestion bias shows that annotators remain capable of reflecting upon the suggested label in general.
Finally, we confirm the quality of the annotated data in transfer learning experiments between different annotator groups.
To facilitate further research in opinion mining on social media data, we release our collected data consisting of 200 expert and 2,785 student annotations.\footnote{Code and data can be found on GitHub: \\ \url{https://github.com/UKPLab/acl2021-label-suggestions-german-covid19}}
\end{abstract}

\section{Introduction}
\label{sec:intro}
The impact analysis of major events like the Covid-19 pandemic is fundamental to research in social sciences. 
To enable more socially sensitive public decision making, researchers need to reliably monitor how various social groups (e.g., political actors, news media, citizens) communicate about political decisions~\citep{jungherr2015analyzing}. 
The increasing use of social media especially allows social science researchers to conduct opinion analysis on a larger scale than with traditional methods, e.g. interviews or questionnaires.
However, the publication of research results is often delayed or temporally transient due to limitations of traditional social science research, i.e. prolonged data gathering processes or opinion surveys being subject to reactivity. 
Given the increasing performance of language models trained on large amounts of data in a self-supervised manner~\citep{devlin-etal-2019-bert,brown2020-language}, one fundamental question that arises is how NLP systems can contribute to alleviate existing difficulties in studies for digital humanities and social sciences~\citep{rischhpidedis2019}.

One important approach to make data annotation more efficient is the use of automated \textit{label suggestions}.
In contrast to \textit{active learning}, that aims to identify a subset of annotated data which leads to optimal model training, label suggestions alleviate the annotation process by providing annotators with pre-annotations (i.e., predictions) from a model~\citep{ringger-etal-2008-assessing, schulz-etal-2019-analysis}.
To enable the annotation of large amounts of data which are used for quantitative analysis by disciplines such as social sciences, label suggestions are a more viable solution than active learning.

One major difficulty with label suggestions is the danger of biasing annotators towards (possibly erroneous) suggestions.
So far, researchers have investigated automated label suggestions for tasks that require domain-specific knowledge~\citep{fort-sagot-2010-influence,yimam-etal-2013-webanno,schulz-etal-2019-analysis}; and have shown that domain experts successfully identify erroneous suggestions and are more robust to potential biases.
However, the limited availability of such expert annotators restricts the use of label suggestions to small, focused annotation studies.
For tasks that do not require domain-specific knowledge and can be conducted with non-expert annotators -- such as crowd workers or citizen science volunteers -- on a large scale, label suggestions have not been considered yet. 
This leads to two open questions.
First, if non-expert annotators that do not receive any training besides annotation guidelines benefit from label suggestions at all.
Second, if existing biases are amplified especially when including interactively updated suggestions that have been shown to be advantageous over static ones~\citep{klie-etal-2020-zero}.

We tackle these challenges by conducting a comparative annotation study with social science students using a recent state-of-the-art model to generate label suggestions~\citep{devlin-etal-2019-bert}.
Our results show that a small set of expert-labeled data is sufficient to improve annotation quality for non-expert annotators. 
In contrast to \citet{schulz-etal-2019-analysis}, we show that although interactive and non-interactive label suggestions substantially improve the agreement, we do not observe significant differences between both approaches.
We further confirm this observation with experiments using models trained on (and transferred to) individual annotator groups.
Our contributions are:
\begin{itemize}
    \item[\textbf{C1}:] An evaluation of label suggestions in terms of annotation quality for non-expert annotators. 
    \item[\textbf{C2}:] An investigation of label suggestion bias for both static and interactively updated suggestions.  
    \item[\textbf{C3}:] A novel corpus of German Twitter posts that can be used by social science researchers to study the effects of governmental measures against Covid-19 on the public opinion.
\end{itemize}

Finally, we also publish 200 expert and 2,785 individual student annotations of our dataset to facilitate further research in this direction.

\section{Related Work}
\label{sec:rw}
\paragraph{Label suggestions.} In an early work, \citet{rehbein-etal-2009-assessing} study the effects of label suggestions on the task of word sense disambiguation and observe a positive effect on annotation quality. 
With the introduction of annotation tools such as brat~\citep{stenetorp-etal-2012-brat}, WebAnno~\citep{yimam-etal-2013-webanno}, or INCEpTION~\citep{klie-etal-2018-inception}, the use of label suggestions became more feasible; leading to an increased investigation of label suggestions in the context of NLP.
For instance, \citet{yimam-etal-2014-automatic} investigate label suggestions for Amharic POS tagging and German named entity recognition and show with expert annotators that label suggestions significantly reduce the annotation time.
Other works further investigate interactively updated label suggestions and come to a similar conclusion~\citep{klie-etal-2020-zero}.
Label suggestions have also been shown to be effective in non-NLP annotation tasks that require domain-specific knowledge such as in medical~\citep{lingren2014evaluating} or educational~\citep{schulz-etal-2019-analysis} use cases.

\paragraph{Bias.}
Annotations from untrained human annotators may introduce biases that are conveyed to machine learning models ~\citep{gururangan-etal-2018-annotation}. 
One possible source of bias may be due to the different decision making process triggered by label suggestions -- namely, first deciding if the suggested label is correct and only if not, considering different labels~\citep{turner2016anchor}. 
Hence, the key question that arises is to what extent annotators are influenced by such suggestions. 
Although \citet{fort-sagot-2010-influence} identify an influence on annotation behaviour when providing pre-annotated data for POS-tagging, they do not measure any clear bias in the annotated labels. 
\citet{rosset-etal-2013-automatic} come to a similar conclusion when investigating the bias introduced by label suggestions in a cross-domain setup, i.e., when using label suggestions from a model that is trained on data from a different domain than the annotated data.
They conduct their experiments with eight annotators from varying levels of expertise and report considerable annotation performance gains while not finding considerable biases introduced by label suggestions. 
Most similar to our work is the setup from \citet{schulz-etal-2019-analysis}.
The authors investigate interactive label suggestions for expert annotators across two domains and study the effects of using existing and newly annotated data for training different suggestion models.
They compare personalised user models against a universal model which has access to all annotated data and show that the latter provides suggestions with a higher acceptance rate.
This seems less surprising due to the substantially larger training set.
Further, they do not identify any bias introduced by pre-annotating data.

Whereas existing work reports no measurable bias for expert annotators~\citep{fort-sagot-2010-influence,lingren2014evaluating,schulz-etal-2019-analysis}, it remains unclear for annotators who have no prior experience in similar annotation tasks; especially for scenarios where -- besides annotation guidelines -- no further training is provided.
However, the use of novice annotators is common for scenarios where no linguistic or domain expertise is required.
Hence, we present a first case-study for the use of interactive label suggestions with non-expert annotators.
Furthermore, we find that recent state-of-the-art models such as BERT~\citep{devlin-etal-2019-bert} can provide high-quality label suggestions with already little training data and hence, are important for interactive label suggestions in non-expert annotation tasks.

\section{Annotation Task}
\label{sec:annottask}
Our task is inspired by social science research on analyzing public opinion using social media~\citep{jungherr2015analyzing,mccormick2017using}.
The goal is to identify opinions in German-speaking countries about governmental measures established to contain the spread of the Corona virus.
We use Twitter due to its international and widespread usage that ensures a sufficient database and the several challenges for the automatic identification of opinions and stance it poses from an NLP perspective~\cite{imran-etal-2016-twitter, mohammad-etal-2016-dataset, gorrell-etal-2019-semeval, conforti-etal-2020-will}.
For example, the use of language varies from colloquial expressions to well-formed arguments and news-spreading statements due to its heterogeneous user base.
Additionally, hashtags are used directly as part of text but also to embed the tweet itself in the broader discussion on the platform.
Finally, the  classification of a tweet is particularly challenging given the character limitation of the platform, i.e., at the date of writing Twitter allows for 280 characters per tweet.

\begin{figure}
    \centering
    \includegraphics[width=0.5\textwidth]{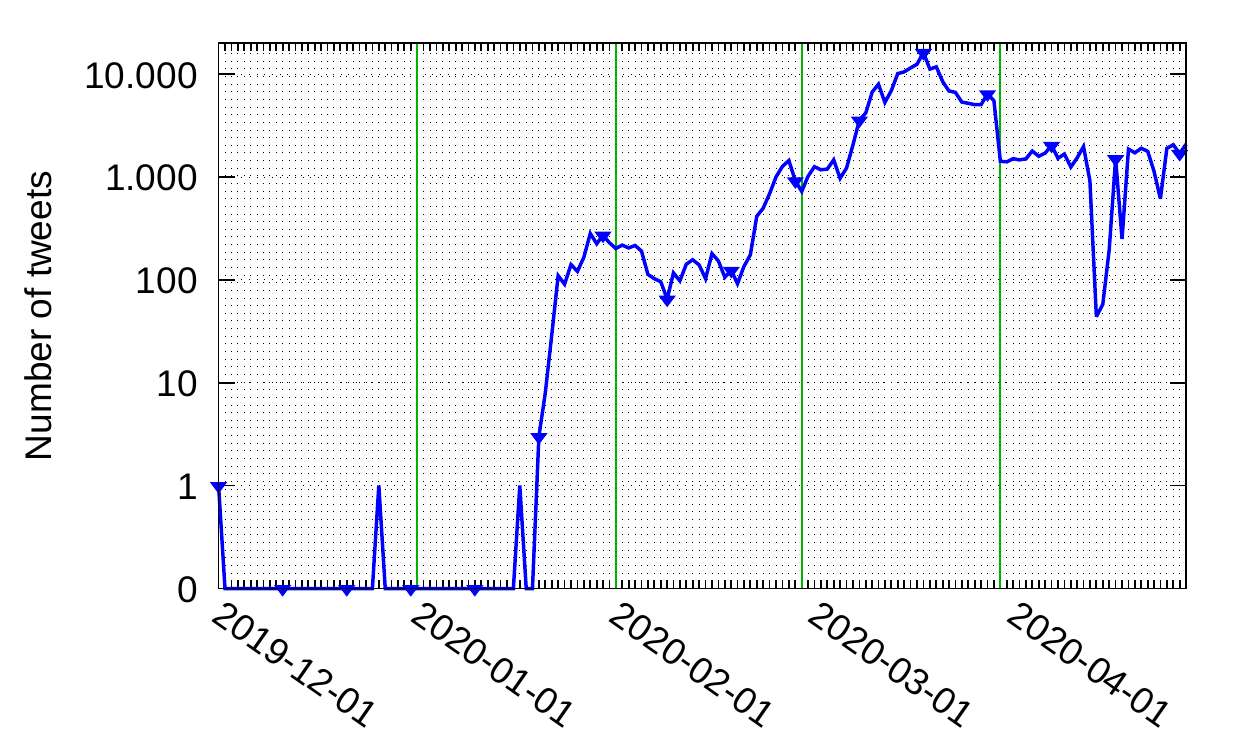}
    \caption{Number of tweets per day collected from December 2019 to April 2020.}
    \label{fig:tweet-time-distribution}
\end{figure}

\paragraph{Data collection.}
Initially, we collected tweets from December 2019 to the end of April 2020.
Using a manually chosen set of search queries (`corona', `pandemie', `covid', `socialdistance'), we made use of the Twitter Streaming API and gathered only those tweets which were classified as German by the Twitter language identifier.
This resulted in a set of approximately 16.5 million tweets.
We retained only tweets that contain key terms referring to measures related to the Covid-19 pandemic and removed all duplicates, retweets and all tweets with text length less than 30 characters.
After filtering, 237,616 tweets remained and their daily temporal distribution is visualized in Figure~\ref{fig:tweet-time-distribution}.
We sample uniformly at random from the remaining tweets for all subsequent annotation tasks.\footnote{We provide additional information about data collection in Appendix~\ref{app:datacrawl} and discuss ethical concerns regarding the use of Twitter data after the conclusion.}

\paragraph{Annotation scheme.} 
We developed annotation guidelines together with three German-speaking researchers from social sciences and iteratively refined them in three successive rounds. 
Our goal from a social science perspective is to analyze the public perception of measures taken by the government.
Therefore, the resulting dataset should help in (1) identifying relevant tweets for governmental measures and if relevant, (2) detecting what stance is expressed.
We follow recent works on stance detection and Twitter data~~\citep{hanselowski-etal-2018-retrospective, baly-etal-2018-integrating, conforti-etal-2020-will} and use four distinct categories for our annotation.
They are defined as follows:
\begin{itemize}[noitemsep,topsep=3pt,itemsep=3pt,itemindent=-1em]
\renewcommand\labelitemi{}    
    \item \textbf{Unrelated}: no measures related to the containment of the pandemic are mentioned
    \item \textbf{Comment}: measures are mentioned, but not assessed or neutral
    \item \textbf{Support}: measures are assessed positively
    \item \textbf{Refute}: measures are assessed negatively
\end{itemize}

The four label annotation scheme allows us to distinguish texts that are related to the pandemic but do not talk about measures (i.e., unrelated).

\begin{figure*}[!htb]
    \centering
    \includegraphics[scale=.5]{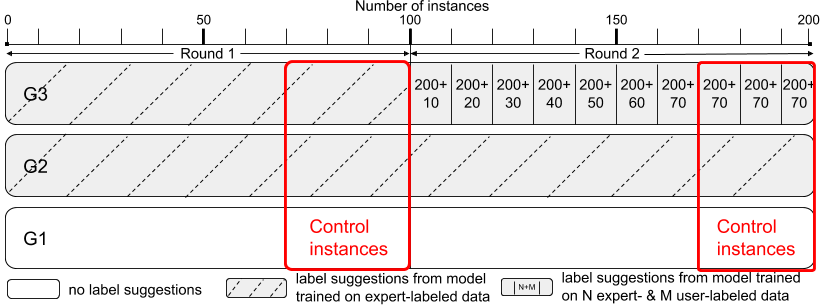}
    \caption{Design of the annotation setup for each of the three user groups. The 30 quality control instances (red) were inserted at random positions but are visualized at the end for presentation purpose.
    }
    \label{fig:annot_setup} 
\end{figure*}

\section{Study Setup}
\label{sec:corpus}
Our goal is to study the effects of interactively updated and static label suggestions in non-expert annotation scenarios.
Non-experts such as crowd workers or student volunteers have no prior experience in annotating comparable tasks and only receive annotation guidelines for preparation.\footnote{We provide the original German guidelines along with the dataset. An English summary is provided in the Appendix~\ref{app:annotation}}
Our secondary goal is to collect a novel dataset that can be used by social science researchers to study the effects of governmental measures for preventing the spread of Covid-19 on the public opinion.

To train a model that provides label suggestions to our non-expert annotators, we first collect a small set of 200 expert-annotated instances.
We then split our non-expert annotators into three different groups that receive (G1) no label suggestions, (G2) suggestions from a model trained on expert annotations, and (G3) suggestions from a model that is retrained interactively using both expert-annotated and interactively annotated data.

\subsection{Expert Annotations}
\label{ssec:expertannotations}
The expert annotations were provided by the researchers (three social science researchers and one NLP researcher) that created the annotation guidelines and who are proficient in solving the task.
In total, 200 tweets were sampled uniformly at random and annotated by all four experts.
The inter-annotator agreement (IAA) across all 200 tweets lies at 0.54 Fleiss's $\kappa$ (moderate agreement) and is comparable to previously reported annotation scores in the field of opinion and argument mining~\cite{bar-haim-etal-2020-arguments, schaefer-stede-2020-annotation, boltuzic-snajder-2014-back}.
Overall, in more than 50\% of the tweets all four experts selected the same label (respectively, in $\sim$75\% of the tweets at least three experts selected the same label). 
The disagreement on the remaining $\sim$25\% of the tweets furthermore shows the increased difficulty of our task due to ambiguities in the data source, e.g., ironical statements or differentiating \textit{governmental} measures from \textit{non-governmental} ones like home-office. 
To compile gold standard labels for instances that the experts disagreed upon, we apply MACE~\cite{hovy-etal-2013-learning} using a threshold of 1.0.
The resulting labels were then re-evaluated by the experts and agreed upon.

\subsection{Student Annotations}

The annotations were conducted with a group of 21 German-speaking university students. 
To ensure a basic level of comparability for our student annotators, we recruited all volunteers from the same social science course at the same university.
The annotators received no further training apart from the annotation guidelines.
We randomly assigned them to three different groups (G1, G2, and G3), each consisting of seven students. 
To investigate the effects of interactive label suggestions, we defined different annotation setups for each group.
The annotations were split into two rounds.
At each round of annotation, students were provided with 100 tweets consisting of 70 new tweets and 30 quality control tweets from the expert-labeled data which are used to compare individual groups.
Across both rounds, we thus obtain a total of 140 unique annotated tweets per student and use 60 tweets for evaluation.
The annotation setup of each group including the individual data splits is visualized in Figure~\ref{fig:annot_setup}.\footnote{Note that the control instances were distributed uniformly at random within a round to mitigate any interdependency effects between different tweets.}

\paragraph{No label suggestions (G1).}
The first group serves as a control group and receives no label suggestions.

\paragraph{Static label suggestions (G2).}
The second group only receives label suggestions based on a model which was trained using the 200 expert-labeled instances described in section~\ref{ssec:expertannotations}.

\paragraph{Interactive label suggestions (G3).}
The last group of students receives expert label suggestions in the first round and interactively updated label suggestions in the second round. 
In contrast to existing work~\citep{schulz-etal-2019-analysis}, this setup allows us to directly quantify effects of bias amplification that may occur with interactive label suggestions.

\subsection{Label Suggestion Model}
\label{ssec:labelsugg}

\paragraph{System setup.}
We conduct our annotation experiments using INCEpTION~\citep{klie-etal-2018-inception} which allows us to integrate label suggestions using recommendation models.
To obtain label suggestions, we use a German version of BERT (\GBERT) that is available through the HuggingFace library~\citep{wolf2019huggingface}.\footnote{\url{https://deepset.ai/german-bert}}
We perform a  random hyperparameter search (cf. Appendix \ref{app:model}) and train the model on the expert annotated data for 10 epochs with a learning rate of 8e-5 and a batch size of 8.
We select the model that performed best in terms of F1-score on a held-out stratified test set (20\% of the data) across ten runs with different random seeds.
All experiments were conducted on a desktop machine with a 6-core 3.8 GHz CPU and a GeForce RTX 2060 GPU (8GB).

\begin{table}[htb]
    \resizebox{\columnwidth}{!}{
    \begin{tabular}{lcc}
    \toprule
    Model & Macro-F1 & Accuracy \\ 
    \midrule
    Majority & .15 & .45 \\
    Random & .23 & .27 \\
    BiLSTM~\cite{schulz-etal-2019-analysis} & .47 & .53 \\
    SBERT+LGBM~\cite{klie-etal-2020-zero} & .50 & .55 \\
    \GBERT (this work) & .66 & .68 \\
    \bottomrule
    \end{tabular}
    }
    \caption{Performance of various label suggestion models on expert-labeled dataset.}
    \label{tab:baseline}
\end{table}

\paragraph{Model comparison.}
To assess the label suggestion quality of our model, we report the predictive performance on the expert-labeled dataset (setup as described above) in Table~\ref{tab:baseline}.
We compare our model with baselines\footnote{We adapted the respective architectures to our setup.} which have been used in related work~\cite{schulz-etal-2019-analysis, klie-etal-2020-zero} for label suggestions.
As expected, \GBERT achieves superior performance and the results are promising for using label suggestions.

\paragraph{Interactive training routine.}
To remedy the cold-start problem, G3 receives label suggestions from the model trained only on the expert-annotated data in round 1.
Afterwards, we retrain the model with an increasing number of instances using both, the expert annotations and the G3 data of individual students from round~1.\footnote{Note that using all previously annotated data of G3 would impair the comparability between individual students as the data was collected asynchronously to allow students to pick their best suited timeslot. Further, a synchronization step between users would impair the applicability of the approach.} 
To avoid unnecessary waiting times for our annotators due to the additional training routine, we always collect batches of 10 instances before re-training our model.
We then repeatedly train individual models for each student in G3 with an increasing amount of data of up to 70 instances.
The 30 expert-annotated quality control tweets were excluded in this step to avoid conflicting labels and duplicated data.

\section{Study Evaluation}
\label{sec:analysis}
\begin{table*}[!htb]
    \centering
    \begin{tabular}{rccrrrr}
    \toprule
    N & Annotator & Avg. Length & \texttt{Unrelated} & \texttt{Comment} & \texttt{Support} & \texttt{Refute} \\ 
    \midrule
    200 & Expert & 189 ($\pm 75$) & 53 (26.5\%) & 89 (44.5\%) & 43 (21.5\%) & 15 (7.5\%) \\
    2,785 & Student & 185 ($\pm 75$) & 1,003 (36.0\%) & 1,055 (37.9\%) & 425 (15.3\%) & 302 (10.8\%) \\
    \midrule
    965 & G1 & 185 ($\pm 76$) & 387 (40.1\%) & 334 (34.6\%) & 128 (13.3\%) & 116 (12.0\%) \\
    980 & G2 & 185 ($\pm 73$) & 320 (32.7\%) & 407 (41.5\%) & 152 (15.5\%) & 101 (10.3\%) \\
    840 & G3 & 184 ($\pm 75$) & 296 (35.2\%) & 314 (37.4\%) & 145 (17.3\%) & 85 (10.1\%)\\
    \bottomrule
    \end{tabular}
    \caption{Our Twitter dataset on public opinion about containment measures during the Corona pandemic.} 
    \label{tab:dataset}
\end{table*}

Table~\ref{tab:dataset} shows the overall statistics of our resulting corpus consisting of 200 expert and 2,785 student-annotated German tweets.
Note that we removed 60 expert-annotated instances that we included for annotation quality control for each student, resulting in 140 annotated tweets per student.

\paragraph{Outliers.} 
A fine-grained analysis of annotation time is not possible due to online annotations at home.
However, one student in G3 had, on average, spent less than a second for each annotation and accepted almost all suggested labels.
This student's annotations were removed from the final dataset and assumed as faulty labels considering the short amount of time spent on this task in comparison to the minimum amount of seven seconds per tweet and annotation for all other students.

\subsection{Annotation Quality}
To assess the overall quality of our collected student annotations, we investigate annotator consistency in terms of inter-annotator-agreement (IAA) as well as the annotator accuracy on our quality assurance instances.

\begin{table}[!htb]
\centering
    \resizebox{\columnwidth}{!}{%
    \begin{tabular}{lcccccc}
    \toprule
    & G1 & & G2 & & G3 & \\ 
    \midrule
    & Acc & IAA & Acc & IAA & Acc & IAA \\
    \midrule
    Round 1 & .74 & .48 & .90 & .76 & .84 & .62\\
    Round 2 & .68 & .47 & .92 & .81 & .82 & .67\\
    Total   & .71 & .48 & .91 & .78 & .83 & .65 \\
    \bottomrule
    \end{tabular}
    }
    \caption{Annotation accuracy (Acc) and IAA (Fleiss' $\kappa$) on the quality control instances for each annotator group and round.}
    \label{tab:accuracy}
\end{table}

Table~\ref{tab:accuracy} shows Fleiss' $\kappa$~\cite{fleiss1971measuring} and the accuracy computed for the quality control instances that were consistent across all groups. 
In general, we observe a similar or higher agreement for our students compared to the expert annotations ($\kappa=0.54$) showing that the guidelines were able to convey the task well. 
We also find that groups that receive label suggestions (G2 and G3) achieve a substantially larger IAA as opposed to G1. 
Most interestingly, we observe a substantial increase in IAA for both G2 and G3 in the second annotation round, whereas the IAA in G1 remains stable.
Analyzing our models' predictions shows that the suggested labels for the 60 quality control samples mostly conform with the label given by the expert (97\% for G2 and 94\% for G3). 
Therefore, annotators are inclined to accept the label suggested by the model. 
We can further confirm this observation when investigating the number of instances that the students labeled correctly (accuracy). 
The highest accuracy is observed for the group that received the highest quality suggestions (G2).
Furthermore, both groups that received label suggestions (G2, G3) express an increased accuracy over the control group (G1). 
In general, for both rounds the accuracy remains similarly high across all groups ($\pm .02$ difference) with only a slight decrease ($-.04$) for G1. 
Hence, we conjecture that the resulting annotations provide satisfying quality given the challenging task and annotator proficiency.

\subsection{Suggestion Bias}
One major challenge in using label suggestions is known in psychology as the \textit{anchoring effect}~\citep{tversky1974judgment, turner2016anchor}.
It describes the concept that annotators who are provided a label suggestion follow a different decision process compared to a group that does not receive any suggestions and tend to accept the suggestions.
As we observe larger IAA and accuracy for groups receiving label suggestions, we look at the label suggestion acceptance rate and which labels have been corrected by the annotators.

\paragraph{Acceptance rate.}
One way to quantify possible biases is to evaluate if annotators tend to accept more suggestions with an increasing number of instances~\citep{schulz-etal-2019-analysis}.
This may be the case when annotators increasingly trust the model with consistently good suggestions.  
Consequently, with increasing trust towards the model's predictions, non-expert annotators may tend to accept more model errors.
To investigate if annotators remain capable of reflecting on instance and label suggestion, we compute the average acceptance rate for G2 and G3 in both rounds.
We find that for both groups, the acceptance rate remains stable (G2: 73\% and 72\%, G3: 68\% and 69\%) and conclude that annotators receiving high quality label suggestions remain critical while producing more consistent results.

\begin{figure}
    \centering
    \includegraphics[width=0.5\textwidth]{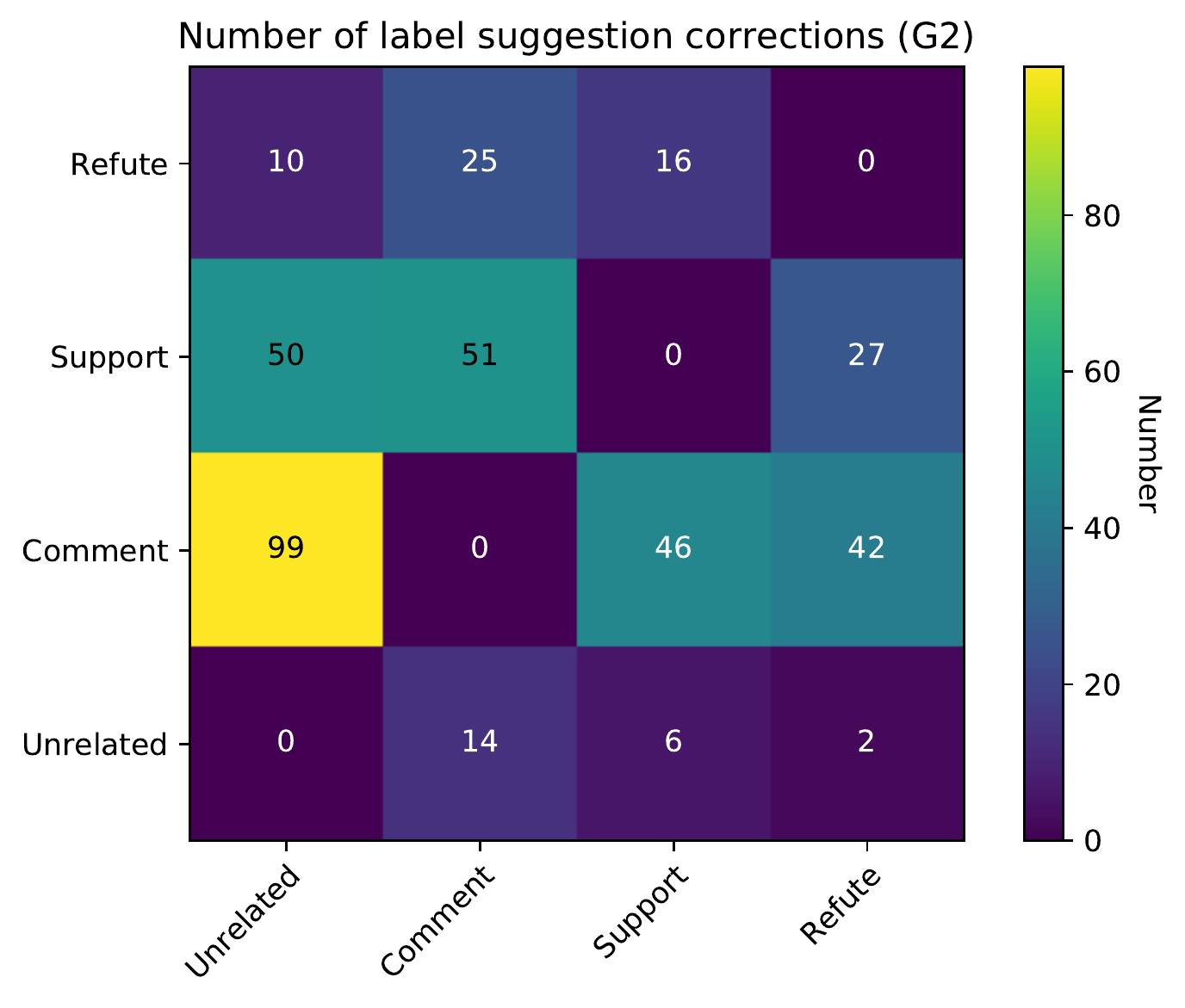}
    \caption{The number of rejected label suggestions. The x-axis displays the corrected label and the y-axis the label suggestion. For example, the upper left corner shows that ten suggestions of label \texttt{Refute} were corrected as \texttt{Unrelated} by the users.}
    \label{fig:g2labelcorrections}
\end{figure}

\begin{figure*}[ht!]
\centering
    \begin{minipage}{0.49\textwidth}
        \centering
          \includegraphics[width=1.0\textwidth]{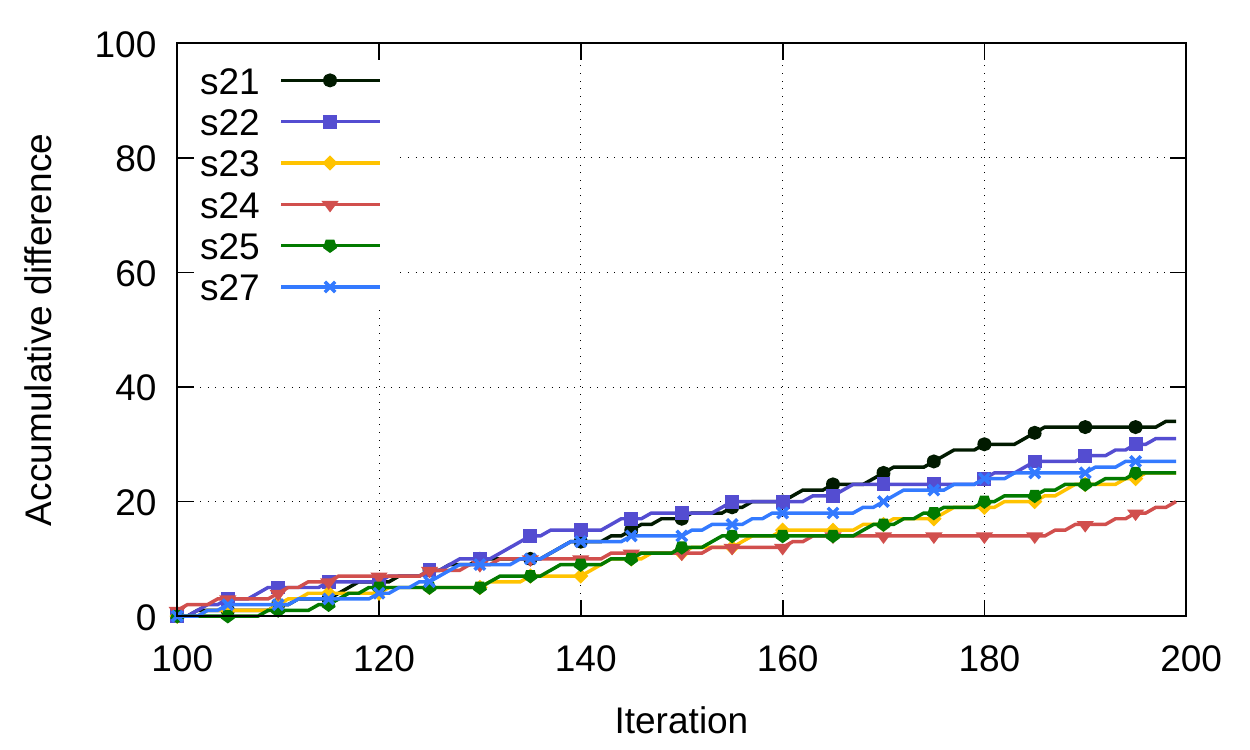}
        \caption{Number of label suggestions that diverge from the model trained on expert-data with increasing number of annotations (for each student).}
        \label{fig:g3-divergence}
    \end{minipage}\hfill 
    \begin{minipage}{0.49\textwidth}
        \centering
        \includegraphics[width=1.0\textwidth]{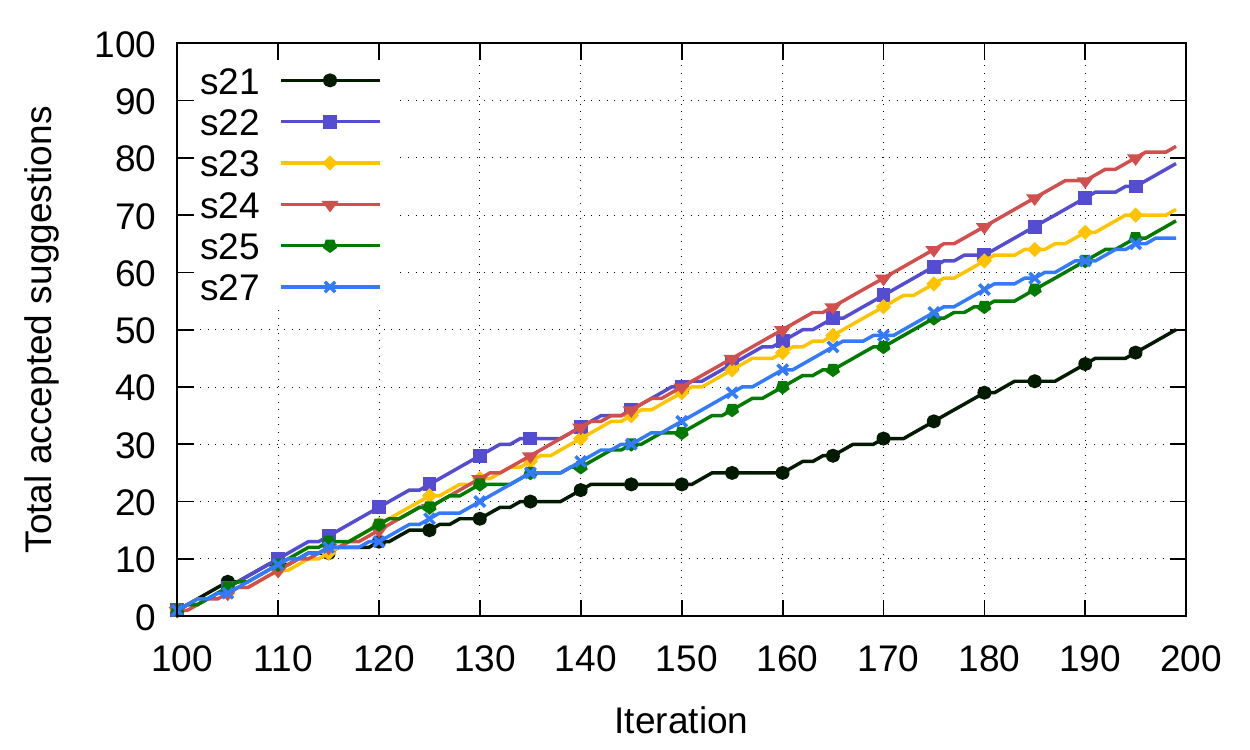}
        \caption{Number of label suggestions in G3 that have been accepted in the second round of annotations (for each student).}
        \label{fig:g3-r2}
    \end{minipage}  
\end{figure*}

\paragraph{Label corrections.}\label{ssec:rejectedsuggestions}
To further evaluate if students are vulnerable to erroneous label suggestions from a model, we specifically investigate labels that have been corrected.
Figure~\ref{fig:g2labelcorrections} shows our results for G2.\footnote{Note that analyzing G3 shows similar observations (cf. Appendix~\ref{app:corrections}).}
As can be seen, the most notable number of label corrections were made by students for unrelated tweets that were classified as comments by the model. 
Additionally, we find a large number of corrections that have been made with respect to the stance of the presented tweet.
We will discuss both types of corrections in the following.

\noindent
\textit{Unrelated tweets.}
The label suggestion model makes the most errors for unrelated tweets (i.e., tweets that are corrected as \texttt{Unrelated}) by mis-classifying them as \texttt{Comment} (99).
In contrast, instances that are identified as \texttt{Unrelated} tweets are only seldomly corrected. 
This indicates an increased focus on recall at the expense of precision for related tweets, most likely due to \texttt{Comment} being the largest class in the training data (see Table~\ref{tab:dataset}, expert data).
We find possible causes for such wrong predictions when we look at examples where \texttt{Comment} was suggested for \texttt{Unrelated} instances\footnote{Note that we present translations of the original German texts for better readability and to protect user privacy}:

\begin{itemize}[noitemsep,topsep=3pt,itemsep=3pt,itemindent=-1em]
\renewcommand\labelitemi{} 
    \item \textbf{Example 1:} The corona virus also requires special protective measures for Naomi Campbell. The top model wears a protective suit during a trip.
    \item \textbf{Example 2:} Extraordinary times call for extraordinary measures: the "Elbschlosskeller" now has a functioning door lock. \#Hamburg \#Corona \#COVID-19
\end{itemize}
\noindent
Clearly, these examples are fairly easy to annotate for humans but are difficult to predict for a model due to specific cue words being mentioned, e.g., \textit{measures}.
Similar results have also been reported in previous work~\cite{hanselowski-etal-2018-retrospective, conforti-etal-2020-will}.
\\\ \\
\noindent
\textit{Stance.}
In Figure~\ref{fig:g2labelcorrections}, we can also see that the model makes mistakes regarding the stance of a tweet.
Especially, 101 \texttt{Support} suggestions have been corrected as either being unrelated or neutral and 88 \texttt{Comment} suggestions have been corrected to either \texttt{Support} or \texttt{Refute}.
For the second case, we often discover tweets that implicitly indicate the stance -- for example, by complaining about people ignoring the measures:

\begin{itemize}[noitemsep,topsep=3pt,itemsep=3pt,itemindent=-1em]
\renewcommand\labelitemi{} 
    \item \textbf{Example 3:} Small tweet aside from XR: Colleague drags himself into the office this morning with flu symptoms (ÖD) The other colleagues still have to convince him to please go home immediately. Only then does he show some understanding. Unbelievable. \#COVID \#SocialDistancing
\end{itemize}
\noindent
Such examples demonstrate the difficulty of the task and seem to be difficult to recognize for the model.
However, given the large amount of label corrections, the non-expert annotators seem to be less susceptible to accept such model errors.

\subsection{Bias Amplification}
The high number of label corrections for specific types of tweets shows that our annotators of G2 remained critical towards the suggested label.
With interactively updated suggestions however, this may not be the case. 
Especially annotators that accept erroneous suggestions may lead to reinforcing a model in its prediction; hence, leading to amplifying biases.

\paragraph{Diverging suggestions.}
To study such effects, we first identify if the interactively updated models express a difference in terms of predictions compared to the static model.
In Figure~\ref{fig:g3-divergence} we can observe that with already 40 instances (Iteration 140), the number of differently predicted instances is ten or higher across all personalized models.
This divergence is highly correlated with the number of changes a student provides (see Figure~\ref{fig:g3-r2}).
We thus can conclude that the interactively trained models are able to adapt to the individual annotations for each annotator.

\begin{figure}[!htb]
    \centering
    \includegraphics[width=0.5\textwidth]{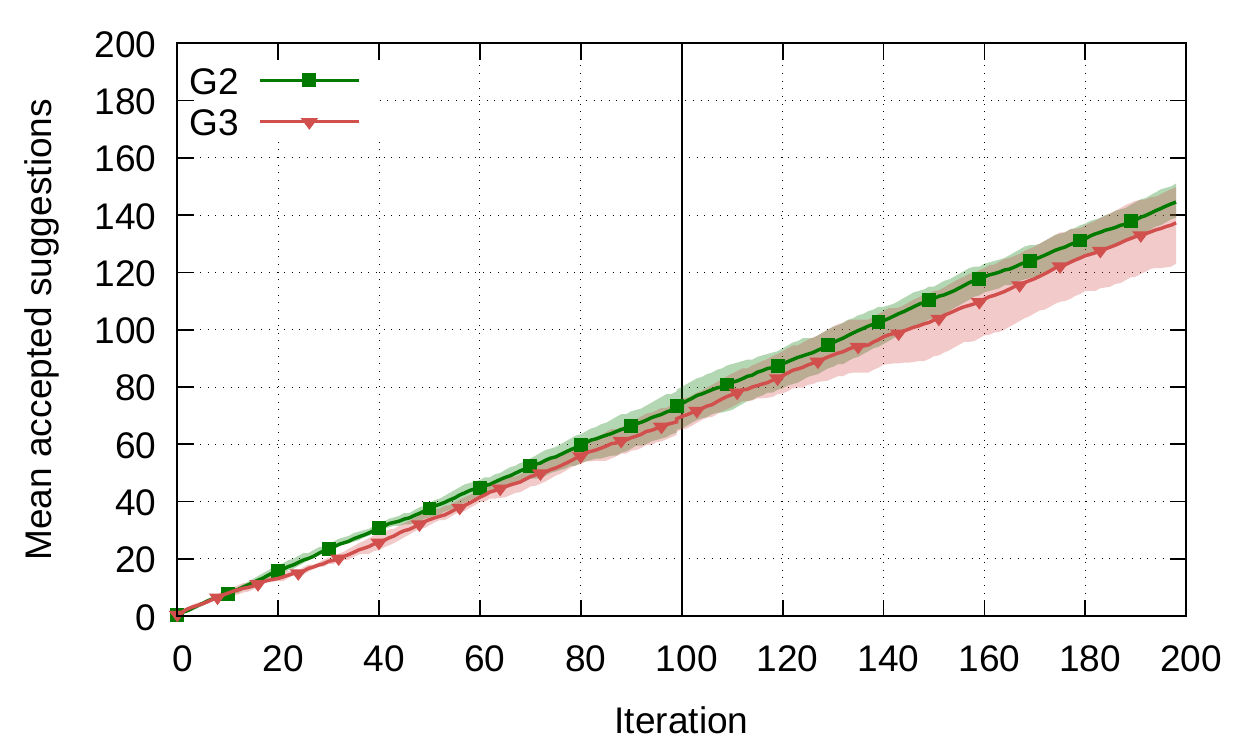}
        \caption{Average number of accepted label suggestions across all instances for G2 and G3. The shared areas display the upper and lower quartile for each group.}
        \label{fig:g2-g3-comparison}
\end{figure}
\paragraph{Comparison to G2.}
Figure~\ref{fig:g2-g3-comparison} shows the average number of accepted suggestions for G2 and G3 as well as the upper and lower quartiles, respectively.
The vertical line separates the first and the second round of annotations.
We find that especially in the first round of annotations, both groups have a very similar acceptance rate of suggested labels.
Only with interactively updated suggestions we find an increasing divergence in G3 with respect to the upper and lower quartiles.

\paragraph{Individual acceptance rate.}
To assess the impact of interactive label suggestions, we further investigate how many suggestions were accepted by each annotator.
Figure~\ref{fig:g3-r2} shows the number of accepted label suggestions for each student in G3 in the second round of annotations. 
Although we observe that the average number of accepted label suggestions remains constant across G2 and G3, we can see substantial differences between individual students. 
For instance, we can observe that for \texttt{s21}, the increased model adaptivity leads to an overall decrease in the number of accepted labels.
Moreover, \texttt{s24} who received predictions that diverge less from the static model prediction accepted the most suggestions in the second round.
This shows that interactive label suggestions does not necessarily lead to a larger acceptance rate -- possibly amplifying biases -- but instead, varies for each annotator and needs to be investigated in future work.

\begin{figure}
    \centering
        \includegraphics[width=0.49\textwidth]{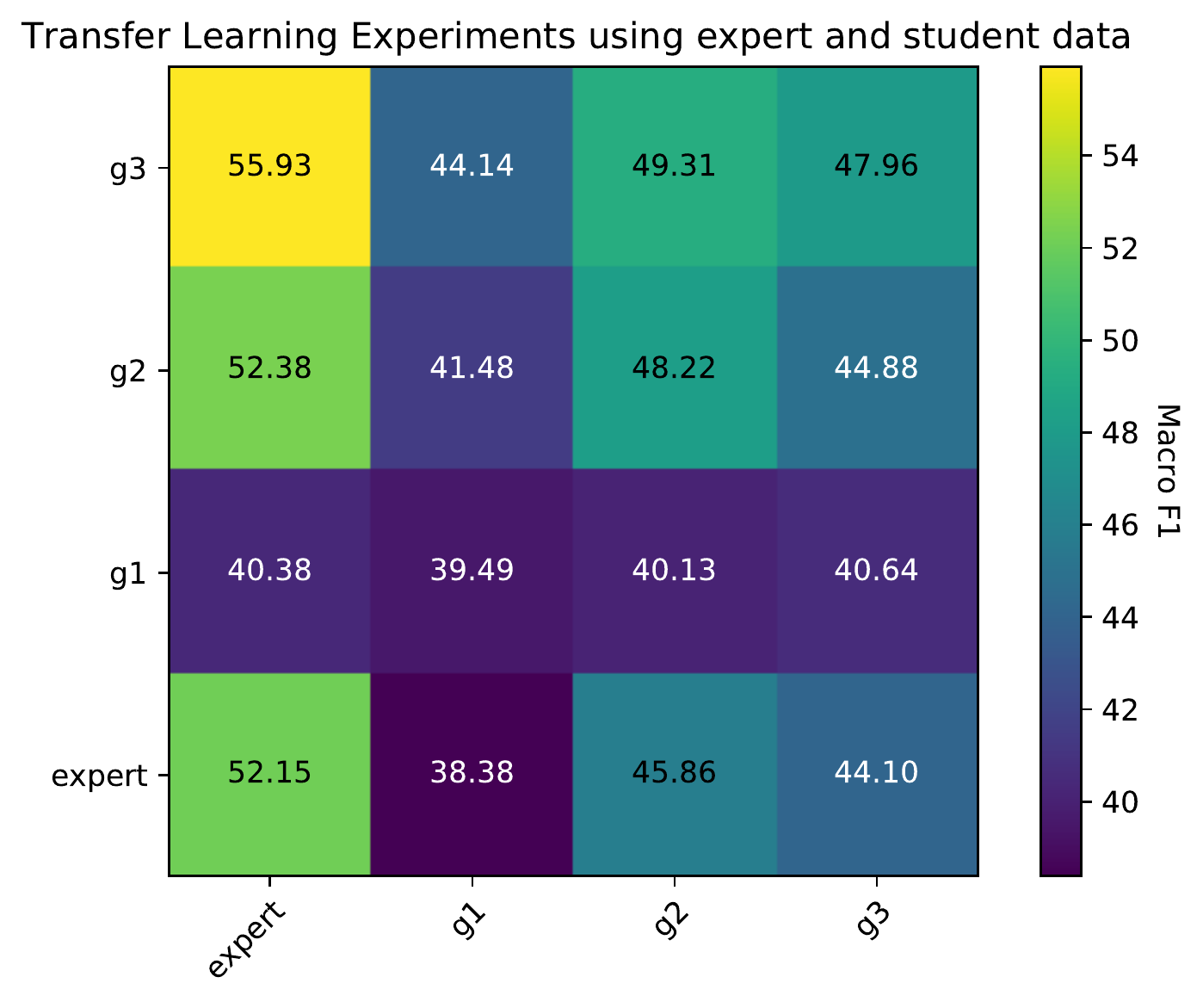}
        \caption{Transfer learning performance of models trained on individual annotator groups. The x-axis presents the dataset which is used for model training, the y-axis lists the dataset used for model testing.}
        \label{fig:transfer}
\end{figure}

\subsection{Cross-group Transfer}

Finally, we investigate how well models trained on different annotator groups transfer to each other.
We hence conduct transfer learning experiments for which we remove the quality control instances in our student groups and train a separate \mbox{\GBERT} model using the same hyperparameters as for the expert model.
We use 80\% of the data for training and the remaining 20\% to identify the best model which we then transfer to another group.
Figure~\ref{fig:transfer} shows the macro-F1 scores averaged across ten independent runs, diagonal entries are the scores on the 20\%.
Most notably, models trained on the groups with label suggestions (G2, G3) do in fact perform comparable or better on the expert-labeled data and outperform models trained on the group not receiving any suggestions (G1).
The higher cross-group performance for models trained on groups that received label suggestions shows that the label suggestions successfully conveyed knowledge from the expert annotated data to our students.

\section{Conclusion}
\label{sec:conclusion}
In this work, we analysed the usefulness of providing label suggestions for untrained annotators to identify opinions in a challenging text domain (i.e., Twitter).
We generated suggestions using expert-labeled training data as well as interactively training models using data annotated by untrained students.
Our results show that label suggestions from a state-of-the-art sentence classification model trained on a small set of expert annotations help improving annotation quality for untrained annotators.
In terms of potential biases that may occur with untrained annotators we observe that the students retained their capability to reflect on the suggested label.
We furthermore do not observe a general amplification in terms of bias with interactively updated suggestions; however, we find that such effects are very specific to individual annotators.
We hence conclude that interactively updated label suggestions need to be considered carefully when applied to non-expert annotation scenarios.

For future work, we plan to leverage our setup to annotate tweets from a larger time span.
In Germany, the measures taken by the government have been met with divided public reaction -- starting with reactions of solidarity and changing towards a more critical public opinion~\citep{viehmann2020b, viehmann2020a}.
In particular, we are interested if our label suggestion model is robust enough to account for such a shift in label distribution.

\section*{Acknowledgements}
This work has been supported by the German Research Foundation (DFG) as part of the Research Training Group KRITIS No. GRK 2222/2.
Further, this work is supported by the European Regional Development Fund (ERDF) and the Hessian State Chancellery – Hessian Minister of Digital Strategy and Development under the promotional reference 20005482 (TexPrax).
We thank the student volunteers from the University of Mainz for their annotations as well as Johannes Daxenberger, Mohsen Mesgar, Ute Winchenbach, Kevin Stowe and the anonymous reviewers for their valuable feedback.

\section*{Ethical considerations}
\label{app:ethics}
\paragraph{Data collection and annotation.}
The tools we use to collect Tweets are in compliance with Twitter's terms of service.
We only release the set of identifiers (Tweet IDs) for the texts used in this research project.
Thereby, we adhere to the Twitter Developer policy\footnote{\url{https://developer.twitter.com/en/developer-terms/agreement-and-policy}} and give users full control of their privacy and data as they can delete or privatize tweets so that they cannot be collected.

We asked student annotators for voluntary participation in the annotation study.
All students have been informed about the goal of the conducted research and the purpose of the collected annotations.
During annotation no information about the tweet's author or any other additional metadata was made available to the annotators.
We did not collect any personal data from the students before, after, or during the annotation task.

\paragraph{Data usage.}
This work presents an investigation of efficient data annotation methods in a case study on social media data.
The results of this work allow social science researchers to apply their analysis on a larger scale. 
In the case of analyzing public opinion on governmental measures, the resulting analysis allows politicians to make more socially sensitive public decisions. 
This information is useful in aggregated form, without the need for information about individual users.
However, we want to point out that users of social media (particularly Twitter) do not constitute a representative sample of the general population, especially in Germany~\cite{newman2020digital}.
Therefore, our goal is not to foster public decision-making solely based upon analysis of Twitter but to provide an additional supporting tool.

\paragraph{Dual use.}
Further, we acknowledge the potential of misuse of our dataset: the annotated data allows anyone, including both individuals and organizations, for training models to identify individuals expressing their consent or dissent with governmental actions.
To this end, we follow the argumentation by \citep{benton-etal-2017-ethical} that in general we cannot prevent publicly available data from being misused but we want to make both researchers and the general public aware of the possible malicious use.

\bibliography{anthology,acl2021}

\begin{thebibliography}{35}
\expandafter\ifx\csname natexlab\endcsname\relax\def\natexlab#1{#1}\fi

\bibitem[{Baly et~al.(2018)Baly, Mohtarami, Glass, M{\`a}rquez, Moschitti, and
  Nakov}]{baly-etal-2018-integrating}
Ramy Baly, Mitra Mohtarami, James Glass, Llu{\'\i}s M{\`a}rquez, Alessandro
  Moschitti, and Preslav Nakov. 2018.
\newblock \href {https://doi.org/10.18653/v1/N18-2004} {Integrating stance
  detection and fact checking in a unified corpus}.
\newblock In \emph{Proceedings of the 2018 Conference of the North {A}merican
  Chapter of the Association for Computational Linguistics: Human Language
  Technologies, Volume 2 (Short Papers)}, pages 21--27, New Orleans, Louisiana.
  Association for Computational Linguistics.

\bibitem[{Bar-Haim et~al.(2020)Bar-Haim, Eden, Friedman, Kantor, Lahav, and
  Slonim}]{bar-haim-etal-2020-arguments}
Roy Bar-Haim, Lilach Eden, Roni Friedman, Yoav Kantor, Dan Lahav, and Noam
  Slonim. 2020.
\newblock \href {https://doi.org/10.18653/v1/2020.acl-main.371} {From arguments
  to key points: {T}owards automatic argument summarization}.
\newblock In \emph{Proceedings of the 58th Annual Meeting of the Association
  for Computational Linguistics}, pages 4029--4039, Online. Association for
  Computational Linguistics.

\bibitem[{Benton et~al.(2017)Benton, Coppersmith, and
  Dredze}]{benton-etal-2017-ethical}
Adrian Benton, Glen Coppersmith, and Mark Dredze. 2017.
\newblock \href {https://doi.org/10.18653/v1/W17-1612} {Ethical research
  protocols for social media health research}.
\newblock In \emph{Proceedings of the First {ACL} Workshop on Ethics in Natural
  Language Processing}, pages 94--102, Valencia, Spain. Association for
  Computational Linguistics.

\bibitem[{Boltu{\v{z}}i{\'c} and
  {\v{S}}najder(2014)}]{boltuzic-snajder-2014-back}
Filip Boltu{\v{z}}i{\'c} and Jan {\v{S}}najder. 2014.
\newblock \href {https://doi.org/10.3115/v1/W14-2107} {Back up your stance:
  Recognizing arguments in online discussions}.
\newblock In \emph{Proceedings of the First Workshop on Argumentation Mining},
  pages 49--58, Baltimore, Maryland. Association for Computational Linguistics.

\bibitem[{Brown et~al.(2020)Brown, Mann, Ryder, Subbiah, Kaplan, Dhariwal,
  Neelakantan, Shyam, Sastry, Askell et~al.}]{brown2020-language}
Tom~B Brown, Benjamin Mann, Nick Ryder, Melanie Subbiah, Jared Kaplan, Prafulla
  Dhariwal, Arvind Neelakantan, Pranav Shyam, Girish Sastry, Amanda Askell,
  et~al. 2020.
\newblock \href
  {https://proceedings.neurips.cc//paper/2020/file/1457c0d6bfcb4967418bfb8ac142f64a-Paper.pdf}
  {Language models are few-shot learners}.
\newblock \emph{arXiv preprint arXiv:2005.14165}.

\bibitem[{Conforti et~al.(2020)Conforti, Berndt, Pilehvar, Giannitsarou,
  Toxvaerd, and Collier}]{conforti-etal-2020-will}
Costanza Conforti, Jakob Berndt, Mohammad~Taher Pilehvar, Chryssi Giannitsarou,
  Flavio Toxvaerd, and Nigel Collier. 2020.
\newblock \href {https://doi.org/10.18653/v1/2020.acl-main.157}
  {Will-they-won{'}t-they: A very large dataset for stance detection on
  {T}witter}.
\newblock In \emph{Proceedings of the 58th Annual Meeting of the Association
  for Computational Linguistics}, pages 1715--1724, Online. Association for
  Computational Linguistics.

\bibitem[{Devlin et~al.(2019)Devlin, Chang, Lee, and
  Toutanova}]{devlin-etal-2019-bert}
Jacob Devlin, Ming-Wei Chang, Kenton Lee, and Kristina Toutanova. 2019.
\newblock \href {https://doi.org/10.18653/v1/N19-1423} {{BERT}: Pre-training of
  deep bidirectional transformers for language understanding}.
\newblock In \emph{Proceedings of the 2019 Conference of the North {A}merican
  Chapter of the Association for Computational Linguistics: Human Language
  Technologies, Volume 1 (Long and Short Papers)}, pages 4171--4186,
  Minneapolis, Minnesota. Association for Computational Linguistics.

\bibitem[{Fleiss(1971)}]{fleiss1971measuring}
Joseph~L Fleiss. 1971.
\newblock \href {https://psycnet.apa.org/buy/1972-05083-001} {Measuring nominal
  scale agreement among many raters.}
\newblock \emph{Psychological bulletin}, 76(5):378.

\bibitem[{Fort and Sagot(2010)}]{fort-sagot-2010-influence}
Kar{\"e}n Fort and Beno{\^\i}t Sagot. 2010.
\newblock \href {https://www.aclweb.org/anthology/W10-1807} {Influence of
  pre-annotation on {POS}-tagged corpus development}.
\newblock In \emph{Proceedings of the Fourth Linguistic Annotation Workshop},
  pages 56--63, Uppsala, Sweden. Association for Computational Linguistics.

\bibitem[{Gorrell et~al.(2019)Gorrell, Kochkina, Liakata, Aker, Zubiaga,
  Bontcheva, and Derczynski}]{gorrell-etal-2019-semeval}
Genevieve Gorrell, Elena Kochkina, Maria Liakata, Ahmet Aker, Arkaitz Zubiaga,
  Kalina Bontcheva, and Leon Derczynski. 2019.
\newblock \href {https://doi.org/10.18653/v1/S19-2147} {{S}em{E}val-2019 task
  7: {R}umour{E}val, determining rumour veracity and support for rumours}.
\newblock In \emph{Proceedings of the 13th International Workshop on Semantic
  Evaluation}, pages 845--854, Minneapolis, Minnesota, USA. Association for
  Computational Linguistics.

\bibitem[{Gururangan et~al.(2018)Gururangan, Swayamdipta, Levy, Schwartz,
  Bowman, and Smith}]{gururangan-etal-2018-annotation}
Suchin Gururangan, Swabha Swayamdipta, Omer Levy, Roy Schwartz, Samuel Bowman,
  and Noah~A. Smith. 2018.
\newblock \href {https://doi.org/10.18653/v1/N18-2017} {Annotation artifacts in
  natural language inference data}.
\newblock In \emph{Proceedings of the 2018 Conference of the North {A}merican
  Chapter of the Association for Computational Linguistics: Human Language
  Technologies, Volume 2 (Short Papers)}, pages 107--112, New Orleans,
  Louisiana. Association for Computational Linguistics.

\bibitem[{Hanselowski et~al.(2018)Hanselowski, PVS, Schiller, Caspelherr,
  Chaudhuri, Meyer, and Gurevych}]{hanselowski-etal-2018-retrospective}
Andreas Hanselowski, Avinesh PVS, Benjamin Schiller, Felix Caspelherr, Debanjan
  Chaudhuri, Christian~M. Meyer, and Iryna Gurevych. 2018.
\newblock \href {https://www.aclweb.org/anthology/C18-1158} {A retrospective
  analysis of the fake news challenge stance-detection task}.
\newblock In \emph{Proceedings of the 27th International Conference on
  Computational Linguistics}, pages 1859--1874, Santa Fe, New Mexico, USA.
  Association for Computational Linguistics.

\bibitem[{Hovy et~al.(2013)Hovy, Berg-Kirkpatrick, Vaswani, and
  Hovy}]{hovy-etal-2013-learning}
Dirk Hovy, Taylor Berg-Kirkpatrick, Ashish Vaswani, and Eduard Hovy. 2013.
\newblock \href {https://www.aclweb.org/anthology/N13-1132} {Learning whom to
  trust with {MACE}}.
\newblock In \emph{Proceedings of the 2013 Conference of the North {A}merican
  Chapter of the Association for Computational Linguistics: Human Language
  Technologies}, pages 1120--1130, Atlanta, Georgia. Association for
  Computational Linguistics.

\bibitem[{Imran et~al.(2016)Imran, Mitra, and
  Castillo}]{imran-etal-2016-twitter}
Muhammad Imran, Prasenjit Mitra, and Carlos Castillo. 2016.
\newblock \href {https://www.aclweb.org/anthology/L16-1259} {{T}witter as a
  lifeline: Human-annotated {T}witter corpora for {NLP} of crisis-related
  messages}.
\newblock In \emph{Proceedings of the Tenth International Conference on
  Language Resources and Evaluation ({LREC}'16)}, pages 1638--1643,
  Portoro{\v{z}}, Slovenia. European Language Resources Association (ELRA).

\bibitem[{Jungherr(2015)}]{jungherr2015analyzing}
Andreas Jungherr. 2015.
\newblock \href {https://doi.org/https://doi.org/10.1007/978-3-319-20319-5}
  {\emph{Analyzing political communication with digital trace data}}.
\newblock Springer, Cham, Switzerland.

\bibitem[{Klie et~al.(2018)Klie, Bugert, Boullosa, Eckart~de Castilho, and
  Gurevych}]{klie-etal-2018-inception}
Jan-Christoph Klie, Michael Bugert, Beto Boullosa, Richard Eckart~de Castilho,
  and Iryna Gurevych. 2018.
\newblock \href {https://www.aclweb.org/anthology/C18-2002} {The {INCE}p{TION}
  platform: Machine-assisted and knowledge-oriented interactive annotation}.
\newblock In \emph{Proceedings of the 27th International Conference on
  Computational Linguistics: System Demonstrations}, pages 5--9, Santa Fe, New
  Mexico. Association for Computational Linguistics.

\bibitem[{Klie et~al.(2020)Klie, Eckart~de Castilho, and
  Gurevych}]{klie-etal-2020-zero}
Jan-Christoph Klie, Richard Eckart~de Castilho, and Iryna Gurevych. 2020.
\newblock \href {https://doi.org/10.18653/v1/2020.acl-main.624} {{F}rom {Z}ero
  to {H}ero: {H}uman-{I}n-{T}he-{L}oop {E}ntity {L}inking in {L}ow {R}esource
  {D}omains}.
\newblock In \emph{Proceedings of the 58th Annual Meeting of the Association
  for Computational Linguistics}, pages 6982--6993, Online. Association for
  Computational Linguistics.

\bibitem[{Lingren et~al.(2014)Lingren, Deleger, Molnar, Zhai, Meinzen-Derr,
  Kaiser, Stoutenborough, Li, and Solti}]{lingren2014evaluating}
Todd Lingren, Louise Deleger, Katalin Molnar, Haijun Zhai, Jareen Meinzen-Derr,
  Megan Kaiser, Laura Stoutenborough, Qi~Li, and Imre Solti. 2014.
\newblock \href {https://doi.org/https://doi.org/10.1136/amiajnl-2013-001837}
  {Evaluating the impact of pre-annotation on annotation speed and potential
  bias: natural language processing gold standard development for clinical
  named entity recognition in clinical trial announcements}.
\newblock \emph{Journal of the American Medical Informatics Association},
  21(3):406--413.

\bibitem[{McCormick et~al.(2017)McCormick, Lee, Cesare, Shojaie, and
  Spiro}]{mccormick2017using}
Tyler~H McCormick, Hedwig Lee, Nina Cesare, Ali Shojaie, and Emma~S Spiro.
  2017.
\newblock \href {https://doi.org/https://doi.org/10.1177\%2F0049124115605339}
  {Using twitter for demographic and social science research: Tools for data
  collection and processing}.
\newblock \emph{Sociological methods \& research}, 46(3):390--421.

\bibitem[{Mohammad et~al.(2016)Mohammad, Kiritchenko, Sobhani, Zhu, and
  Cherry}]{mohammad-etal-2016-dataset}
Saif Mohammad, Svetlana Kiritchenko, Parinaz Sobhani, Xiaodan Zhu, and Colin
  Cherry. 2016.
\newblock \href {https://www.aclweb.org/anthology/L16-1623} {A dataset for
  detecting stance in tweets}.
\newblock In \emph{Proceedings of the Tenth International Conference on
  Language Resources and Evaluation ({LREC}'16)}, pages 3945--3952,
  Portoro{\v{z}}, Slovenia. European Language Resources Association (ELRA).

\bibitem[{Newman et~al.(2020)Newman, Fletcher, Schulz, Andi, and
  Nielsen}]{newman2020digital}
Nic Newman, Richard Fletcher, Anne Schulz, Simge Andi, and Rasmus-Kleis
  Nielsen. 2020.
\newblock \href
  {https://reutersinstitute.politics.ox.ac.uk/sites/default/files/2020-06/DNR_2020_FINAL.pdf}
  {Digital news report 2020}.
\newblock \emph{Reuters Institute for the Study of Journalism}, pages 2020--06.

\bibitem[{Rehbein et~al.(2009)Rehbein, Ruppenhofer, and
  Sporleder}]{rehbein-etal-2009-assessing}
Ines Rehbein, Josef Ruppenhofer, and Caroline Sporleder. 2009.
\newblock \href {https://www.aclweb.org/anthology/W09-3003} {Assessing the
  benefits of partial automatic pre-labeling for frame-semantic annotation}.
\newblock In \emph{Proceedings of the Third Linguistic Annotation Workshop
  ({LAW} {III})}, pages 19--26, Suntec, Singapore. Association for
  Computational Linguistics.

\bibitem[{Ringger et~al.(2008)Ringger, Carmen, Haertel, Seppi, Lonsdale,
  McClanahan, Carroll, and Ellison}]{ringger-etal-2008-assessing}
Eric Ringger, Marc Carmen, Robbie Haertel, Kevin Seppi, Deryle Lonsdale, Peter
  McClanahan, James Carroll, and Noel Ellison. 2008.
\newblock \href
  {http://www.lrec-conf.org/proceedings/lrec2008/pdf/832_paper.pdf} {Assessing
  the costs of machine-assisted corpus annotation through a user study}.
\newblock In \emph{Proceedings of the Sixth International Conference on
  Language Resources and Evaluation ({LREC}'08)}, Marrakech, Morocco. European
  Language Resources Association (ELRA).

\bibitem[{Risch et~al.(2019)Risch, Stoll, Ziegele, and
  Krestel}]{rischhpidedis2019}
Julian Risch, Anke Stoll, Marc Ziegele, and Ralf Krestel. 2019.
\newblock \href
  {https://corpora.linguistik.uni-erlangen.de/data/konvens/proceedings/papers/germeval/Germeval_Task_2_2019_paper_10.HPIDEDIS.pdf}
  {hpidedis at germeval 2019: Offensive language identification using a german
  bert model}.
\newblock In \emph{Proceedings of the 15th Conference on Natural Language
  Processing (KONVENS 2019)}, pages 405--410, Erlangen, Germany. German Society
  for Computational Linguistics \& Language Technology.

\bibitem[{Rosset et~al.(2013)Rosset, Grouin, Lavergne, Ben~Jannet, Leixa,
  Galibert, and Zweigenbaum}]{rosset-etal-2013-automatic}
Sophie Rosset, Cyril Grouin, Thomas Lavergne, Mohamed Ben~Jannet,
  J{\'e}r{\'e}my Leixa, Olivier Galibert, and Pierre Zweigenbaum. 2013.
\newblock \href {https://www.aclweb.org/anthology/W13-2321} {Automatic named
  entity pre-annotation for out-of-domain human annotation}.
\newblock In \emph{Proceedings of the 7th Linguistic Annotation Workshop and
  Interoperability with Discourse}, pages 168--177, Sofia, Bulgaria.
  Association for Computational Linguistics.

\bibitem[{Schaefer and Stede(2020)}]{schaefer-stede-2020-annotation}
Robin Schaefer and Manfred Stede. 2020.
\newblock \href {https://www.aclweb.org/anthology/2020.argmining-1.6}
  {Annotation and detection of arguments in tweets}.
\newblock In \emph{Proceedings of the 7th Workshop on Argument Mining}, pages
  53--58, Online. Association for Computational Linguistics.

\bibitem[{Schulz et~al.(2019)Schulz, Meyer, Kiesewetter, Sailer, Bauer,
  Fischer, Fischer, and Gurevych}]{schulz-etal-2019-analysis}
Claudia Schulz, Christian~M. Meyer, Jan Kiesewetter, Michael Sailer, Elisabeth
  Bauer, Martin~R. Fischer, Frank Fischer, and Iryna Gurevych. 2019.
\newblock \href {https://doi.org/10.18653/v1/P19-1265} {Analysis of automatic
  annotation suggestions for hard discourse-level tasks in expert domains}.
\newblock In \emph{Proceedings of the 57th Annual Meeting of the Association
  for Computational Linguistics}, pages 2761--2772, Florence, Italy.
  Association for Computational Linguistics.

\bibitem[{Stenetorp et~al.(2012)Stenetorp, Pyysalo, Topi{\'c}, Ohta, Ananiadou,
  and Tsujii}]{stenetorp-etal-2012-brat}
Pontus Stenetorp, Sampo Pyysalo, Goran Topi{\'c}, Tomoko Ohta, Sophia
  Ananiadou, and Jun{'}ichi Tsujii. 2012.
\newblock \href {https://www.aclweb.org/anthology/E12-2021} {brat: a web-based
  tool for {NLP}-assisted text annotation}.
\newblock In \emph{Proceedings of the Demonstrations at the 13th Conference of
  the {E}uropean Chapter of the Association for Computational Linguistics},
  pages 102--107, Avignon, France. Association for Computational Linguistics.

\bibitem[{Turner and Schley(2016)}]{turner2016anchor}
Brandon~M Turner and Dan~R Schley. 2016.
\newblock \href
  {https://doi.org/https://doi.org/10.1016/j.cogpsych.2016.07.003} {The anchor
  integration model: A descriptive model of anchoring effects}.
\newblock \emph{Cognitive Psychology}, 90:1--47.

\bibitem[{Tversky and Kahneman(1974)}]{tversky1974judgment}
Amos Tversky and Daniel Kahneman. 1974.
\newblock \href {https://doi.org/https://doi.org/10.1126/science.185.4157.1124}
  {Judgment under uncertainty: Heuristics and biases}.
\newblock \emph{science}, 185(4157):1124--1131.

\bibitem[{Viehmann et~al.(2020{\natexlab{a}})Viehmann, Ziegele, and
  Quiring}]{viehmann2020b}
Christina Viehmann, Marc Ziegele, and Oliver Quiring. 2020{\natexlab{a}}.
\newblock \href
  {https://www.ard-werbung.de/media-perspektiven/fachzeitschrift/artikel/detailseite-2020/gut-informiert-durch-die-pandemie-nutzung-unterschiedlicher-informationsquellen-in-der-corona-krise/}
  {Gut informiert durch die {P}andemie? {N}utzung unterschiedlicher
  {I}nformationsquellen in der {C}orona-{K}rise. {E}rgebnisse einer
  dreiwelligen {P}anelbefragung im {J}ahr 2020}.
\newblock \emph{Media Perspektiven}, 10-11:2556--577.

\bibitem[{Viehmann et~al.(2020{\natexlab{b}})Viehmann, Ziegele, and
  Quiring}]{viehmann2020a}
Christina Viehmann, Marc Ziegele, and Oliver Quiring. 2020{\natexlab{b}}.
\newblock \href
  {https://doi.org/https://www.faz-institut.de/wp-content/uploads/sites/2/2020/06/Viehmann-k_2020_02-GESCHUETZT.pdf}
  {In der {K}rise r{\"u}cken alle etwas mehr zusammen? {F}ast! – {W}ie
  {M}edien und andere {F}aktoren zur gef{\"u}hlten {I}ntegration und {S}paltung
  beitragen}.
\newblock \emph{Kommunikationsmanager}, 2:32--36.

\bibitem[{Wolf et~al.(2020)Wolf, Debut, Sanh, Chaumond, Delangue, Moi, Cistac,
  Rault, Louf, Funtowicz, Davison, Shleifer, von Platen, Ma, Jernite, Plu, Xu,
  Scao, Gugger, Drame, Lhoest, and Rush}]{wolf2019huggingface}
Thomas Wolf, Lysandre Debut, Victor Sanh, Julien Chaumond, Clement Delangue,
  Anthony Moi, Pierric Cistac, Tim Rault, Rémi Louf, Morgan Funtowicz, Joe
  Davison, Sam Shleifer, Patrick von Platen, Clara Ma, Yacine Jernite, Julien
  Plu, Canwen Xu, Teven~Le Scao, Sylvain Gugger, Mariama Drame, Quentin Lhoest,
  and Alexander~M. Rush. 2020.
\newblock \href {https://www.aclweb.org/anthology/2020.emnlp-demos.6}
  {Transformers: State-of-the-art natural language processing}.
\newblock In \emph{Proceedings of the 2020 Conference on Empirical Methods in
  Natural Language Processing: System Demonstrations}, pages 38--45, Online.
  Association for Computational Linguistics.

\bibitem[{Yimam et~al.(2014)Yimam, Biemann, Eckart~de Castilho, and
  Gurevych}]{yimam-etal-2014-automatic}
Seid~Muhie Yimam, Chris Biemann, Richard Eckart~de Castilho, and Iryna
  Gurevych. 2014.
\newblock \href {https://doi.org/10.3115/v1/P14-5016} {Automatic annotation
  suggestions and custom annotation layers in {W}eb{A}nno}.
\newblock In \emph{Proceedings of 52nd Annual Meeting of the Association for
  Computational Linguistics: System Demonstrations}, pages 91--96, Baltimore,
  Maryland. Association for Computational Linguistics.

\bibitem[{Yimam et~al.(2013)Yimam, Gurevych, Eckart~de Castilho, and
  Biemann}]{yimam-etal-2013-webanno}
Seid~Muhie Yimam, Iryna Gurevych, Richard Eckart~de Castilho, and Chris
  Biemann. 2013.
\newblock \href {https://www.aclweb.org/anthology/P13-4001} {{W}eb{A}nno: A
  flexible, web-based and visually supported system for distributed
  annotations}.
\newblock In \emph{Proceedings of the 51st Annual Meeting of the Association
  for Computational Linguistics: System Demonstrations}, pages 1--6, Sofia,
  Bulgaria. Association for Computational Linguistics.

\end{thebibliography}
\bibliographystyle{acl_natbib}

\clearpage

\appendix

\section{Data Crawl Details}
\label{app:datacrawl}

\subsection{Filter Terms}
We crawled Twitter posts using Live Streaming API (Twitter).
Based on a preliminary examination of the data, we selected the following set of terms for filtering tweets:
\texttt{['stayhomesavelifes', 'wirbleibenzuhause', 'bleibdaheim', 'abstandhalten', 'flatthecurve', 'flattenthecurve', 'sperre', 'verbot', 'beschraenkung', 'quarantäne', 'quarantaene', 'wirvsvirus', 'schließung', 'homeoffice', 'infektionsschutz', 'ansteckungsrisiko', 'notbetrieb', 'bleibtzuhause', 'stayhome']} 

\subsection{Additional Data}
Unlike related work~\citep{schaefer-stede-2020-annotation}, we do not investigate reply structures of tweets. In preliminary experiments we found that our collection method provides a large enough amount of relevant tweets which can be annotated without context.

\section{Annotation}
\label{app:annotation}

In this section we provide more detailed information on the annotation guidelines and the annotation platform we used.

\subsection{Annotation Guidelines}

We first provide background context about the measures for containing the spread of Covid-19.
Afterwards, we provide a definition for measures to be considered during the annotation study:
\textit{We consider all measures which are taken by the government to contain the pandemic (e.g., closing of schools)}.

Next, we introduce the annotation task as a two-step process. 
First, the annotator has to decide if the text actually does mention measures as defined as above (some examples are provided for clarification).
If not, the annotator selects the label \texttt{NoMeasure}\footnote{Please note that we use a different notation in the main paper. The label \texttt{NoMeasure} corresponds to \texttt{Unrelated}, label \texttt{NoOpinion} corresponds to \texttt{Comment}, \texttt{ProOpinion} corresponds to \texttt{Support} and \texttt{ConOpinion} corresponds to \texttt{Refute}}.
In the opposite case, the annotator decides in a second step if the text contains a positive position (\texttt{ProOpinion}), a negative position (\texttt{ConOpinion}) or if there is no stance expressed (\texttt{NoOpinion}).
We provide examples for each label to our annotators.

During our preliminary studies, we identified several ambiguities regarding the stance annotation which is in the nature of the source of the texts (Twitter) and the subject of the annotation (measures regarding the Covid-19 pandemic):
\begin{itemize}
    \item \textbf{a Tweet discusses (positive/negative) consequences or by-products of measures} : we regard those as (positive/negative) statements as the author implicitly states their opinion by reflecting upon the measures
    \item \textbf{a Tweet reflects the opinion of another actor}: this is considered as an opinion as defined above. It is assumed that the author posts this opinion because they identify themselves with the original opinion.
    \item \textbf{a Tweet makes an unagitated observation whether measures are functioning}: this is not to be taken as an opinion for or against the measures per se. Only if an explicit assessment of the observation is made, the position can be derived.
    \item \textbf{the role of Hashtags}: Hashtags are often ambiguous and the respective context needs to be taken into account. Therefore, in our annotation hashtags are only considered as context to what is said; they never stand for themselves. Hashtags can be used to determine whether a measure is being addressed. To do this, the hashtag must contain a measure. Further, hashtags can be used as context to support the position in a tweet.
\end{itemize}

These decisions are reflected at the corresponding positions in the annotation guidelines, along with several example tweets.
In the end we provide a note that Twitter posts may contain malicious, suggestive,
offensive, or potentially sensitive content and that the
annotation can be paused and resumed at any time.

\subsection{Annotation Interface}

\begin{figure*}
 \centering 
 \includegraphics[scale=0.33]{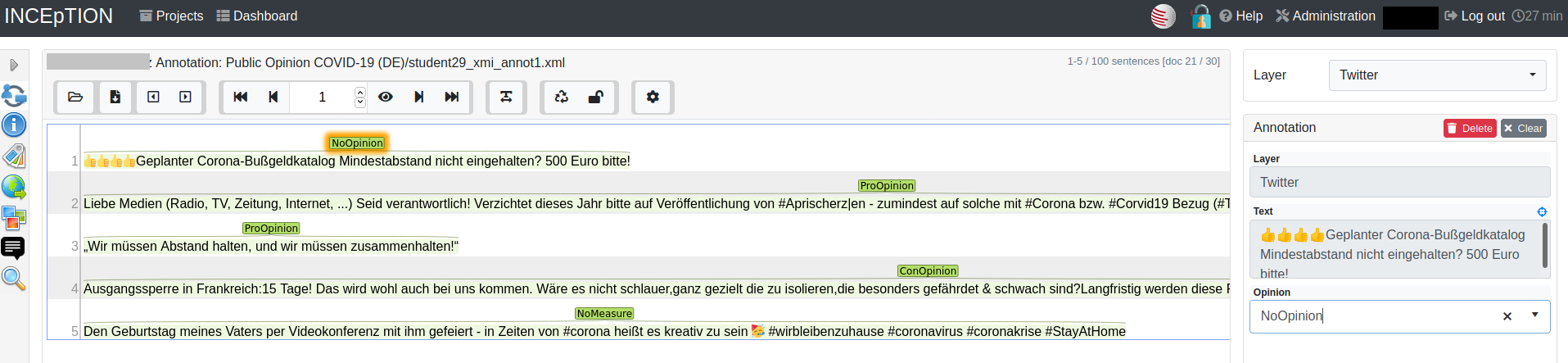}
 \caption{A screenshot of the annotation interface using INCEpTION~\cite{klie-etal-2018-inception}}
 \label{fig:annot_interface}
\end{figure*}

In Figure~\ref{fig:annot_interface} a screenshot of the annotation interface is depicted.
It is taken from the group were label recommendations are provided.
The Twitter posts to be annotated are shown in the center where each line corresponds to a single tweet.
For the sake of clarity, only five texts are shown simultaneously and the user navigates through all texts using the navigation bar above the text window.

The label recommendations are displayed using a green box above the corresponding text and the currently selected recommendation is highlighted in orange.
If the user agrees with the provided label, nothing needs to be changed.
In the opposite case, the user can click on the recommendation and select another label on the right-hand side (\texttt{Annotation} panel) using the \texttt{Opinion} dropdown field.
The annotators receiving no label suggestions (G1) do not see any recommendation during annotation.
They create an annotation for each sentence by double-clicking on the sentence.
Once the user has finished annotating all samples, the annotation session is finished by clicking the lock symbol in the navigation bar.
The technical procedure of the annotation has been explained to all annotators beforehand.

\subsection{Label Suggestion Model}
\label{app:model}
We used the \texttt{german-bert-cased} BERT base model which was pretrained on a German Wikipedia Dump (6GB), an OpenLegalData dump (2.4GB) and news articles (3.6GB).
It was trained for 810k steps with a batch size of 1024 for sequence length 128 and 30k steps with sequence length 512.
It outperformed the multilingual version of BERT on several downstream tasks using German data (GermEval-2018\footnote{\url{https://projects.fzai.h-da.de/iggsa/germeval-2018/}}, GermEval-2014 NER\footnote{\url{https://sites.google.com/site/germeval2014ner/data}}, 10kGNAD\footnote{\url{https://tblock.github.io/10kGNAD/}}).
More information can be found at the corresponding website\footnote{\url{https://deepset.ai/german-bert}}.

For our setup, we performed a random hyperparameter search using the following combinations:
\begin{itemize}
    \item learning rate: [0.01, 0.1, 0.001, 0.0001, 0.00001, 0.00005, 0.00008]
    \item batch size: [4,8,16]
\end{itemize}

\section{Label Rejections}
\label{app:corrections}

\begin{figure}[!htb]
    \centering
    \includegraphics[width=0.5\textwidth]{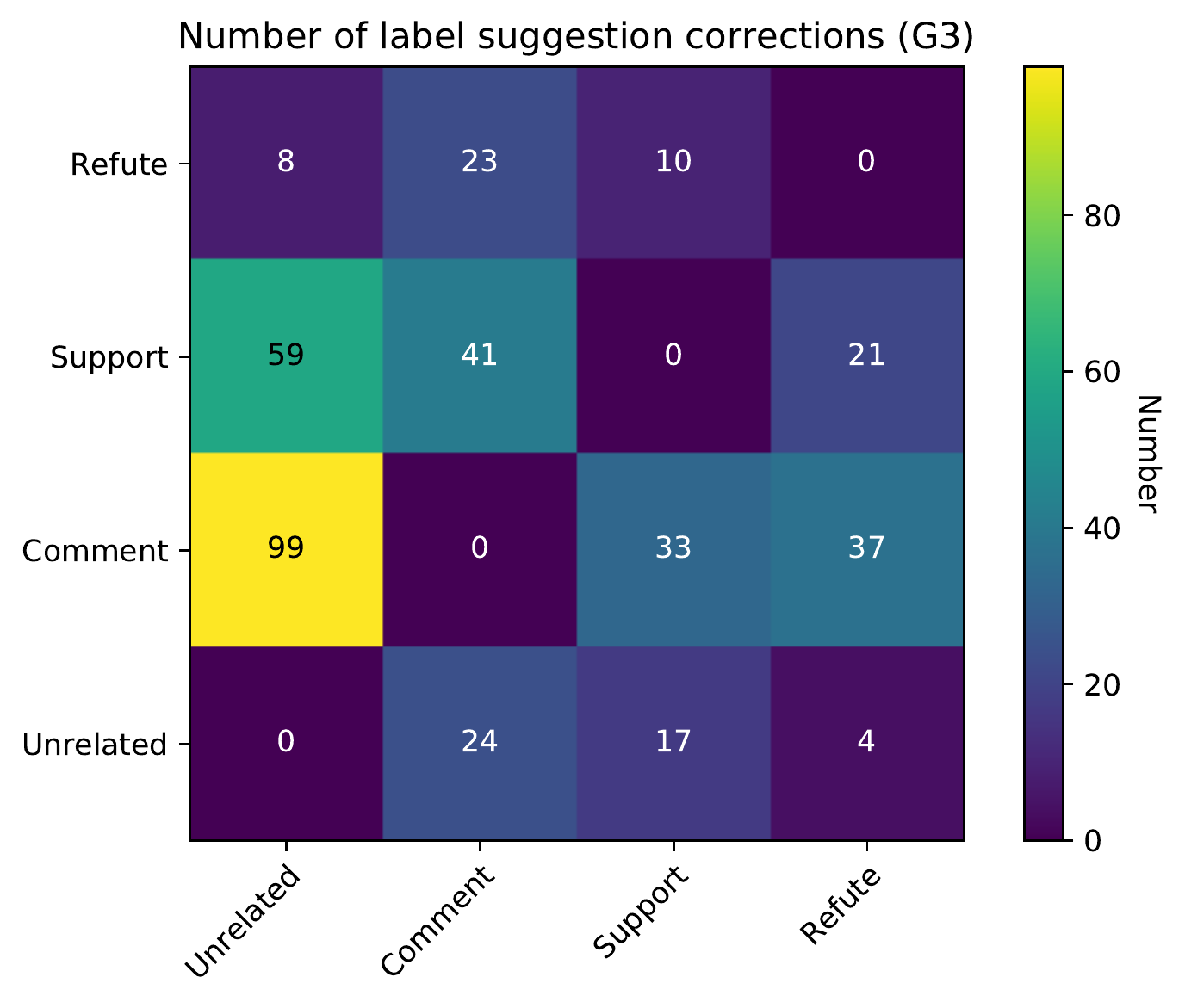}
    \caption{Number of rejected label suggestions for group G3. The x-axis displays the corrected label and the y-axis the label suggestion. For example, the upper left corner shows that 8 suggestions of label \texttt{Refute} were corrected as \texttt{Unrelated} by the users.}
    \label{fig:g3labelcorrections}
\end{figure}

Figure~\ref{fig:g3labelcorrections} displays how student annotators from G3 corrected label suggestions, per category.
As discussed in Section~\ref{ssec:rejectedsuggestions} we observe a similar pattern as for annotator group G2.
The majority of label corrections are for the predicted category \texttt{Comment} or corrections for a wrongly predicted stance (e.g., predictions of \texttt{Support} or \texttt{Refute}).

\end{document}